\documentclass[conference]{IEEEtran}
\usepackage{url}
\usepackage{cite}
\usepackage{amsmath,amssymb,amsfonts}
\usepackage{algorithmic}
\usepackage[hidelinks]{hyperref}
\usepackage{graphicx}
\usepackage{textcomp}
\usepackage{xcolor}
\usepackage{tikz}
\usetikzlibrary{positioning}
\usepackage[acronym]{glossaries}
\usepackage{subcaption}
\usepackage{stfloats}
\usepackage{soul}
\def\BibTeX{{\rm B\kern-.05em{\sc i\kern-.025em b}\kern-.08em
    T\kern-.1667em\lower.7ex\hbox{E}\kern-.125emX}}

\makeatletter
\newcommand{\newlineauthors}{%
  \end{@IEEEauthorhalign}\hfill\mbox{}\par
  \mbox{}\hfill\begin{@IEEEauthorhalign}
}
\makeatother    

\newcommand\etal[1]{#1 \emph{et al.}}
    
\newacronym{ann}{ANN}{Artificial Neural Network} 
\newacronym{snn}{SNN}{Spiking Neural Network}    
\newacronym{dnn}{DNN}{Deep Neural Network}
\newacronym{ac}{AC}{Accumulated Computation} 
\newacronym{mac}{MAC}{Multiply-Accumulate Computation} 
\newacronym{lif}{LIF}{Leaky Integrate-and-Fire}
\newacronym{plif}{PLIF}{Parametric Leaky Integrate-and-Fire} 
\newacronym{bn}{BN}{Batch Normalization} 
\newacronym{slayer}{SLAYER}{Spike Layer Error Reassignment in Time}
\newacronym{rnn}{RNN}{Recurrent Neural Network}
\newacronym{bptt}{BPTT}{Backpropagation Through Time}
\newacronym{tbptt}{tBPTT}{truncated Backpropagation Through Time}
\newacronym{kd}{KD}{Knowledge Distillation}
\newacronym{map}{mAP}{mean Average Precision}
\newacronym{gen1}{GEN1}{GEN1 Automotive Detection}
\newacronym{nms}{NMS}{Non-Maximum Suppression}

\begin{document}

\title{Spiking CenterNet: A Distillation-boosted Spiking Neural Network for Object Detection}

\author{\IEEEauthorblockN{Lennard Bodden}
\IEEEauthorblockA{\textit{Department NetMedia} \\
\textit{Fraunhofer IAIS}\\
St. Augustin, Germany \\
lennard.bodden@iais.fraunhofer.de}
\and
\IEEEauthorblockN{Duc Bach Ha\textsuperscript{*}}
\IEEEauthorblockA{\textit{Department NetMedia} \\
\textit{Fraunhofer IAIS}\\
St. Augustin, Germany \\
duc.bach.ha@iais.fraunhofer.de}
\and
\IEEEauthorblockN{Franziska Schwaiger\textsuperscript{*}}
\IEEEauthorblockA{\textit{Department Dependable Perception and Imaging} \\
\textit{Fraunhofer IKS}\\
Munich, Germany \\
franziska.schwaiger@iks.fraunhofer.de}
\newlineauthors
\IEEEauthorblockN{Lars Kreuzberg}
\IEEEauthorblockA{\textit{Department NetMedia} \\
\textit{Fraunhofer IAIS}\\
St. Augustin, Germany \\
lars.kreuzberg@iais.fraunhofer.de}
\and
\IEEEauthorblockN{Sven Behnke}
\IEEEauthorblockA{\textit{Computer Science Institute VI} \\
\textit{University of Bonn}\\
Bonn, Germany \\
behnke@cs.uni-bonn.de}
}

\maketitle
\begin{abstract}
In the era of AI at the edge, self-driving cars, and climate change, the need for energy-efficient, small, embedded AI is growing. 
Spiking Neural Networks (SNNs) are a promising approach to address this challenge, with their event-driven information flow and sparse activations.
We propose Spiking CenterNet for object detection on event data.
It combines an SNN CenterNet adaptation with an efficient M2U-Net-based decoder.
Our model significantly outperforms comparable previous work on Prophesee's challenging GEN1 Automotive Detection Dataset while using less than half the energy. 
Distilling the knowledge of a non-spiking teacher into our SNN further increases performance.
To the best of our knowledge, our work is the first approach that takes advantage of knowledge distillation in the field of spiking object detection.

\end{abstract}


\begingroup\renewcommand\thefootnote{*}
\footnotetext{Equal contribution}

\section{Introduction}

In recent years, the integration of object detection capabilities into edge devices has witnessed unprecedented growth, driven by the ever-increasing demand for real-time applications in fields such as automotive and robotics. Edge devices, characterized by their resource-constrained nature, pose unique challenges in terms of computational efficiency and power consumption. Addressing these challenges requires innovative approaches that not only provide accurate object detection but also ensure power-efficiency for operation.

\begin{figure*}[htbp]
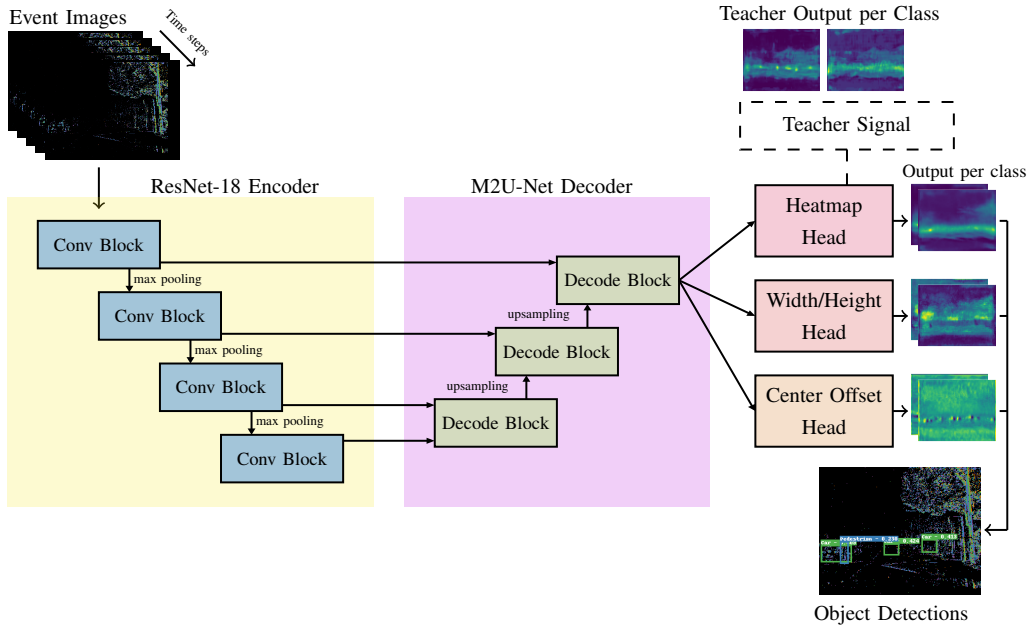

    \centering
    \include{architecture}\vspace*{-6ex}
    \caption{Overview of our spiking object detection model. We combine a ResNet-18 encoder with M2U-Net-based decoding \cite{laibacher2019m2unet} to feed into CenterNet-based heads \cite{zhou2019centernet}. We remove any residual connections, and replace all activation functions with \gls{plif} neurons. Postprocessing calculates bounding boxes from the head output.}
    \label{fig:architecture}
\end{figure*}

One promising approach for achieving these goals is the utilization of \glspl{snn}, which are inspired by the communication mechanism of biological neurons. \glspl{snn} exhibit inherent power-efficiency as their distinctive feature is event-driven information processing, achieved through all-or-nothing events (spikes) for communication between neurons. This attribute sets \glspl{snn} apart from conventional \glspl{ann}, which primarily rely on non-binary floating-point values (\textit{floats}). Spikes facilitate fast, cost-effective neuron interactions via single-bit electronic impulses, unlike multi-bit data like floats which demand multiple impulses. In addition, sparse binary value transmission conserves energy by keeping most of the neurons inactive during operation.

Similarly to \glspl{snn}, event-based cameras also exhibit asynchronous behavior, aligning well with \glspl{snn}' processing capabilities. Event-based cameras provide several benefits compared to conventional frame-based cameras: they have an exceptional temporal resolution in microseconds \cite{Gallego2020}, rendering them ideal for applications demanding real-time responsiveness. Furthermore, they excel in energy efficiency, transmitting data only in response to sensory input changes instead of transmitting redundant information as conventional frame-based cameras do. \glspl{snn} and event-based cameras work together effectively, providing fast, energy-efficient data processing.

The development of \gls{snn}-based object detectors holds substantial promise for advancing the utilization of \glspl{snn} in real-time autonomous applications demanding energy-efficient object detection capabilities, unlike the predominant focus of previous \gls{snn} research on classification tasks. However, the effective employment of \glspl{snn} remains a significant challenge due to the intrinsic difficulty associated with directly training these networks, given their discrete and spiking (i.e., binary) nature and thus non-differentiable activations. 

In this work, we propose a novel fully \gls{snn}-based object detection framework (see Fig. \ref{fig:architecture}) trained on automotive data recorded by event-based cameras. Our key contributions are as follows: 

\begin{itemize}
    \item We propose a modified, spiking version of the simple and versatile CenterNet architecture \cite{zhou2019centernet} which is also - to the best of our knowledge - the first trained \gls{snn} detector that does not require costly \gls{nms}. 
    \item We replace CenterNet's upsampling by the more efficient modules from M2U-Net  \cite{laibacher2019m2unet} and add binary skip connections between encoder and decoder which improves gradient flow despite the spiking communication. 
    \item To the best of our knowledge, we are the first that utilize \gls{kd} for \glspl{snn} in the context of object detection, with the aim of addressing the challenges associated with training efficiency and model generalization.
\end{itemize}

Our \gls{snn}-based object detector outperforms comparable previous work on the challenging \gls{gen1} dataset by 2.6\,\% \gls{map}. We show the effectiveness of \gls{kd}, which improves model performance in terms of mAP by an average of 1.8\,\% over a baseline \gls{snn}. We also show that our model achieves better power efficiency than its non-spiking counterpart and the state-of-the-art \gls{snn}-based object detectors.


\section{Related Work}
\subsection{Learning with Spiking Neural Networks}
At the individual neuron level, there are many different neuron models to use, ranging from the very detailed Hodgkin and Huxley model used in neurobiology \cite{hodgkin1952quantitative} to the much simpler and because of this very popular \gls{lif} model \cite{delorme1999spikenet}. As a good trade-off between complexity and efficiency, we choose the \gls{plif} neuron \cite{fang2020incorporating}, which is a \gls{lif} with a learnable membrane variable.

However, all these neuron models share the disadvantages of training complexity and the spike output's lower information resolution.
Firstly, due to the discrete and non-differentiable nature of spike-based activations, the common practice of  back-propagating errors through the network during training cannot be performed directly. 
Secondly, the temporal aspect turns \glspl{snn} into a type of \gls{rnn}, which are inherently difficult to train \cite{Pascanu2013}.
Finally, a series of binary spikes with practical length can only represent a limited amount of different values in contrast to the high precision of floating points.  

 Due to the aforementioned non-differentiable nature of spikes, special frameworks are needed to enable training. Some widely-used methods are \gls{slayer} \cite{shrestha2018slayer} and surrogate gradient learning \cite{neftci2019surrogate}.
While \gls{slayer} uses a temporal credit assignment policy to back-propagate errors to previous layers, surrogate gradients simply approximate the non-differentiable spiking function with a similar differentiable function. We choose surrogate gradients since they enable treating an \gls{snn} as a simple \gls{rnn}, which allows utilization of established learning algorithms such as \gls{bptt} \cite{rumelhart1987learning}. 

\subsection{Spiking Object Detectors}
There are currently two main directions for implementing spiking object detectors: conversion and training from scratch. Converting the weights of a usually isomorphic non-spiking \gls{ann} is popular for creating complex \glspl{snn} because it avoids training non-differentiable spiking functions. However, these conversions often result in loss of accuracy, which is why the bulk of work in this direction goes into minimizing conversion loss. SpikingYolo \cite{kim2020spiking} is an example of a successful conversion from \gls{ann} to \gls{snn} for object detection. However, this network requires at least 1\,000 time steps to detect objects with acceptable accuracy. 
Recently, \cite{qu2023spiking} achieved good accuracy with only four time steps, but they stretch the definition of \glspl{snn}. They use non-binary "burst spikes" and spike weighting, which is used to make spike signals more complex, but they disregard the computational impact of it. While conversion enables the reuse of an existing well-trained network, the high number of time steps or more complex spikes both negatively affect the resulting network's efficiency.
Furthermore, conversion from a non-recurrent \gls{ann} does not allow the resulting \gls{snn} to take advantage of temporal event data.

An alternative approach is training \glspl{snn} from scratch. Cordone et al.\cite{cordone2022object} introduced the first fully spiking \gls{snn} for object detection trained on a challenging real-world event dataset. This was an important milestone for \gls{snn} research as it showed the feasibility of training from scratch and a low- time step \gls{snn}. Cordone followed up with \cite{cordone2022performance} in which a new, continuous training setup with \gls{tbptt} is successfully employed on 60s sequences of event data. Unlike the previous work, the resulting \gls{snn} creates predictions each time step. Su et al. \cite{su2023deep} developed an even better performing SNN model by introducing a "spiking residual block". However, it includes non-spiking residual connections which violate the \gls{snn}'s core principle of spiking signals between layers. We also opt to train our \gls{snn} object detector from scratch to fully utilize the inherent sparsity of trained \glspl{snn} and  improve upon the previous fully-spiking standard set by \cite{cordone2022object}. Furthermore, we utilize the better performance of the non-spiking counterpart as a teacher signal for our \gls{snn} model through \gls{kd}. 

\subsection{Knowledge Distillation for SNNs}
The idea of Knowledge Distillation (KD) is a well-established learning strategy introduced by Hinton et al.~\cite{hinton2015distilling}.
KD is about improving the performance of a smaller, more efficient "student" network by transferring the knowledge of a larger, more capable "teacher" network as an additional soft learning target to the student network.
First examples of using \gls{kd} for \glspl{snn} are limited to simple classification problems.
While Tran et al.~\cite{tran2022training} use a more traditional \gls{kd} approach together with \gls{snn}-to-\gls{ann} conversion, Xu et al.~\cite{xu2023biologically} use a novel approach they call "re-\gls{kd}" in which they adapt the network structure on-the-fly while distilling knowledge. Our approach, described in Section \ref{sec:kd}, is closer to the former. 

\gls{kd} in object detection~\cite{chen2017learning} is more challenging than in classification , though. In classification, the precise values of the output layer matter little; it is only required that the output neuron associated with the correct class fires most strongly. Therefore, a teacher signal produced by non-spiking \gls{ann} can be easily mirrored by an \gls{snn} by increasing incoming spikes at the correct output neuron. However, the output of object detection models is not only a lot more complex, but some form of regression is also often required to produce detailed bounding box coordinates. We address this issue by choosing a network in which object localization is disentangled from the bounding box regression and thus enables easy \gls{kd} of an object \textit{heatmap} from a non-spiking teacher. There is -- to the best of our knowledge -- no previous work of \gls{kd} with \gls{snn}-based object detectors.

\section{Method}

\subsection{Spiking CenterNet}
The main motivation behind constructing our \gls{snn} architecture is simplicity, as we find that complex neural network structures, albeit proven for non-spiking \glspl{ann}, function worse with spiking activations. For example, highly optimized and complex architectures such as EfficientDet \cite{tan2020efficientdet} suffer from the binarization of feature maps and contain modules such as a singular global feature factor which as a spike may disable a module's output completely. 
Therefore, we construct our model from two very simple architectures: CenterNet \cite{zhou2019centernet} and M2U-Net \cite{laibacher2019m2unet}.

Due to its simplicity and reproducibility, CenterNet \cite{zhou2019centernet} has become a very influential object detection model. 
It features variable backbones and heads for different tasks which encompass 2D and 3D bounding box detection as well as human pose estimation. 
The key idea of the model is to estimate objects or target points (e.g., joints) as key points (activity blobs centered at target) on 2D classification heatmaps (one for each class, cf. Fig.~\ref{fig:hm_output}). 
These heatmaps divide the input image into a grid of variable size. 
To balance the grid's coarseness, an offset regression with similar shape is also produced. 
Additionally, depending on the task, bounding box width/height regression may also be used. 
These predictions are made by individual heads which take feature maps of roughly the same size as the input. This makes the backbone structure similar to a segmentation network with an encoder and a decoder part.  
 
The final bounding box predictions are produced by an extraction of local maxima from the heatmap, which replaces the typical \gls{nms} \cite{bodla2017soft} found in other detection models. This allows us to fully utilize the time dimension and produce several outputs with spiking CenterNet heads with a hidden spiking layer of 64 channels for the heatmap, offset regression and bounding box width/height regression \cite{zhou2019centernet}, rather than only aggregating features over time and performing a single-step detection as done by \cite{cordone2022object}. In our work, we take the mean of each head's output over all five time steps to produce a final, more robust output. This allows the model to be independent of the specific number of time steps and generate results with fewer time steps if needed (see Fig. \ref{fig:ablation}). 

Among the different backbones, we opt for ResNet-18 \cite{he2016deepb} due to its simplicity and relatively small size. We adopt the SpikingJelly's implementation \cite{spikingjelly} of the network and replace the classic ReLU activation with SpikingJelly's \gls{plif} neuron throughout the network, including the decoder. We replace the first convolutional layer to adapt to the number of input channels depending on the data (i.e., 4 channels for event data, see Section \ref{subsubsec:data}).

\subsection{M2U-Net Decoding} \label{subsec:m2unet}
M2U-Net \cite{laibacher2019m2unet} is a popular small segmentation network with 0.55M parameters. It features an encoder-decoder structure similar to the ResNet-18 based backbone in CenterNet \cite{zhou2019centernet}. However, M2U-Net uses a static upsampling step rather than weight-based deconvolution as in CenterNet's decoding.

Originally, CenterNet's ResNet-18-based version uses so-called \textit{transposed convolutions} or \textit{deconvolutions} \cite{zeiler2010deconvolutional} to increase the feature maps' size before feeding them to CenterNet's heads. However, this relies heavily on the ability of the network to compress spatial information in the rather low-resolution, but high-dimensional feature maps of the encoding steps. It also requires the network to learn meaningful deconvolutional weights. Since \glspl{snn} are quite limited in feature map output due to their binary nature and generally work better with fewer tunable weights, we instead choose a decoding strategy based on M2U-Net.

\begin{figure}
    \centering
    \tikzset{every picture/.style={line width=0.75pt}} 
\newcommand{\leftx}{-2.5}
\newcommand{\rightx}{2.5}

\begin{tikzpicture}[yscale=-1,xscale=1]

\tikzstyle{every node}=[rounded corners=0.25cm, font=\scriptsize]
\node[draw, align=center] (input) at (\leftx,-1){\textbf{Input}};

\node[draw, align=center, fill={rgb:yellow,1;white,1}] (pw) at (\leftx, 0){
\textbf{Pointwise}\\Conv2d (in, hidden, k=1)\\ BatchNorm \\ Relu6};

\node[draw, align=center, fill={rgb:yellow,1;white,1}] (dw) at (\leftx, 1.5){
\textbf{Depthwise}\\Conv2d (hidden, hidden, k=3)\\ BatchNorm \\ Relu6};

\node[draw, align=center, fill={rgb:yellow,1;white,1}] (pw_lin) at (\leftx, 3){
\textbf{Pointwise linear}\\Conv2d (hidden, out, k=1)\\ BatchNorm};

\node[draw, circle, align=center] (add) at (\leftx, 4){+};

\node[draw, align=center] (output) at (\leftx,5){\textbf{Output}};

\draw [->](input) -- (pw);
\draw [->] (pw) -- (dw);
\draw [->] (dw) -- (pw_lin);
\draw [->] (pw_lin) -- (add);
\draw (input) -- (\leftx + 2, -1);
\draw (\leftx +2, -1) -- (\leftx + 2, 4);
\draw [->] (\leftx + 2, 4) -- (add);
\draw [->] (add) -- (output);

\node[draw, align=center] (input) at (\rightx,-1){\textbf{Input}};

\node[draw, align=center, fill={rgb:blue,1;white,3}] (pw) at (\rightx, 0){
\textbf{Pointwise}\\Conv2d (in, hidden, k=1)\\ BatchNorm \\ PLIF};

\node[draw, align=center, fill={rgb:blue,1;white,3}] (dw) at (\rightx, 1.5){
\textbf{Depthwise}\\Conv2d (hidden, hidden, k=3)\\ BatchNorm};

\node[draw, align=center, fill={rgb:blue,1;white,3}] (pw_lin) at (\rightx, 3){
\textbf{Pointwise linear}\\Conv2d (hidden, out, k=1)\\ BatchNorm \\ PLIF};

\draw[red, thick, dotted] (\rightx-2, 0.8) rectangle (\rightx +2, 3.8);
\node[rotate=90, font=\scriptsize] (m) at (\rightx + 1.8, 2.4){merged for inference};

\node[draw, align=center] (output) at (\rightx,5){\textbf{Output}};

\draw [->] (input) -- (pw);
\draw [->] (pw) -- (dw);
\draw [->] (dw) -- (pw_lin);
\draw [->] (pw_lin) -- (output);

\node[align=center, font=\small] (a) at (\leftx, 5.7){\textbf{Inverted Residual}};
\node[align=center, font=\small] (a) at (\rightx, 5.7){\textbf{Spiking Expansion}};

\draw[dashed] (0, -1.5) -- (0, 5.8);

\end{tikzpicture}
    \caption{Differences between M2UNet's \cite{laibacher2019m2unet} original \textit{Inverted Residual} block and our spiking adaptation which drops the non-binary residual connection and moves the activation function from the depth-wise to the point-wise linear block. These two can be merged during inference, thus making the entire block fully spiking.}
    \label{fig:spiking_block}
\end{figure}
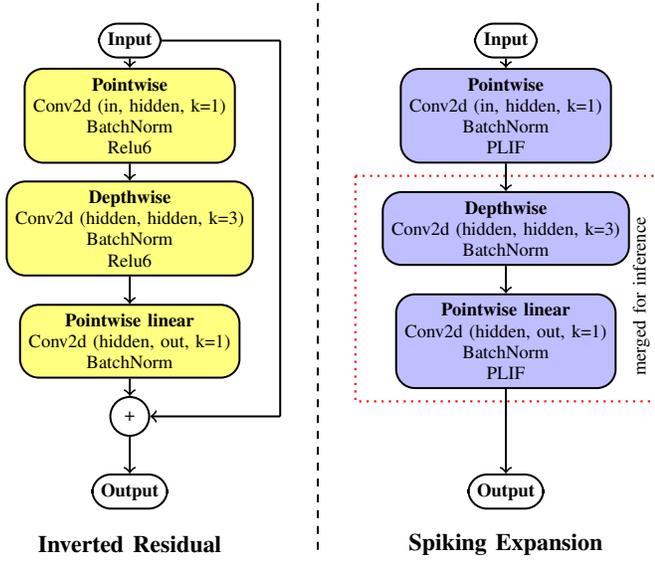

M2U-Net's decoding \cite{laibacher2019m2unet} is particularly suited for \glspl{snn}. Its skip connections between encoder and decoder allow the \gls{snn} to retain important high-level information more easily. Moreover, unlike additive residual connections the concatenation of these skip connections does not violate the binary nature of the \gls{snn}. Furthermore, M2UNet's simple weightless upsampling spares the \gls{snn} from unnecessary weights. We add M2U-Net's upsampling blocks and connect the encoder blocks' outputs with the  decoder blocks of the same input size (see Fig. \ref{fig:architecture}). We replace the ReLU activations with \gls{plif} neurons. However, in M2U-Net's \textit{Inverted Residual Block}, we drop the activation function between the depth-wise and the point-wise linear convolutions so we can merge them for inference, thus eliminating non-spike signals between these layers (see Fig. \ref{fig:spiking_block}). Furthermore, we remove the identity connection in the Inverted Residual connections, as the summation of the identity and residual and the resulting non-binary values violates the idea of a (binary) \gls{snn}. We find that a Boolean OR-operation as a binary alternative does not improve the result, and instead, we decide to drop the identity connection entirely.

In this way, we create a combination of two networks that we call Spiking CenterNet, which offers a flexible arrangement consisting of simple, \gls{snn}-compatible building blocks.

\subsection{Knowledge Distillation for SNNs} \label{sec:kd}
As the isomorphic, non-spiking counterpart of our \gls{snn} model performs better than the \gls{snn}, we try to distill knowledge from this non-spiking version to the spiking one. 
Our approach is straightforward:  
First, the non-spiking teacher is trained separately and then the weights are frozen during the training of the \gls{snn}. For each time step during the latter, we pass the same input to both the \gls{snn} and the teacher. Finally, the teacher's output is used as a soft target signal to calculate the mean squared error:
\begin{equation}
    L_\text{teach} = \frac{1}{T}\sum_{t=0}^{T-1} \sum_{p \in \text{Pixels}} \lbrace o_p(t) - \hat{o}_p(t)\rbrace^2 ,
\end{equation}
where $o_p(t)$ is the output of the $p$-th pixel of the \gls{snn} model at time $t$ and $\hat{o}_p(t)$ is the corresponding teacher output. With this we arrive at an overall loss of:
\begin{equation}
    L = L_{CN} + \alpha \cdot L_\text{teach},
\end{equation}
where $L_{CN}$ is the CenterNet loss as in \cite{zhou2019centernet}. We choose $\alpha=1$ after initial experiments.

\subsection{Measuring Energy Consumption of ANNs and SNNs}

One of the most important benefits of \glspl{snn}---compared to \glspl{ann}---is their energy efficiency. 
Measuring this advantage, however, is nontrivial as suitable \gls{snn} hardware is not yet available. 
One of the key factors for measuring energy consumption is the number of synaptic operations in a network. In the case of \glspl{ann}, this is marked by a \gls{mac}, multiplying each non-spiking activation with the respective weight before adding it to the internal sum. For \glspl{snn}, however, the synaptic operation only consists of a cheaper \gls{ac}, because only incoming spikes in the binary activations are processed without any multiplication. 
However, many modern \glspl{snn} trade efficient energy consumption for more accuracy by also using non-spike operations that entail \glspl{mac}. According to \cite{chen2023training}, an approximation for the energy consumption can thus be determined by the number of \gls{ac} and \gls{mac} operations:
\begin{equation}
    E_\text{SNN} = T \cdot (f \cdot E_\text{AC} \cdot O_\text{AC} + E_\text{MAC} \cdot O_\text{MAC}).
\end{equation}

Here, T is the simulation time and $f$ is the average firing rate. $E_\text{AC}, E_\text{MAC}$ and $O_\text{AC}, O_\text{MAC}$ are the energy consumption and number of operations for \gls{ac} and \gls{mac}, respectively. We assume energy consumption values of $E_\text{AC}=0.9\text{pJ}$ and $E_\text{MAC}=4.6\text{pJ}$ based on current 45\,nm technology following related works (\cite{horowitz2014computing}, \cite{su2023deep}).

This rough estimation of energy consumption enables a fair comparison between our work and others such as \cite{cordone2022object} and \cite{su2023deep}, which also report \glspl{mac} and \glspl{ac}.
Other works like \cite{dampfhoffer2022snns} and \cite{lemaire2022analytical} also try to consider memory as another important factor. However, this requires assumptions on the memory usage strategy, which diversely affects energy consumption, and goes beyond the scope of this work.  


\section{Experiments}

\begin{table*}[ht]
 \centering \small
 \caption{Results on the \gls{gen1} dataset~\cite{tournemire2020prophesee}.}
    \newcommand{\g}[1]{\textcolor{gray}{#1}}
    \newcommand{\horspace}{.3cm}
{   
    \begin{tabular}{l|c|c c c|c|c}
        \hline
          & \textbf{\#Params} & \textbf{mAP} &&& \textbf{Time steps} & \textbf{Energy/time step}\\
         \textbf{Model} & (Millions)& best & avg & std && (mJ)  \\ \hline \hline
         Non-spiking \glspl{ann}: &&&&&\\
         \hspace{\horspace} HMNet-L3 \cite{hamaguchi2023hierarchical} & - & 0.471 & - & - & - & - \\
         \hspace{\horspace} Ours (ANN teacher) & 12.97 & 0.278 & 0.275 & 0.0044 & 1 & 28.214\\ \hline
         \glspl{snn}: &&&&& \\
         \hspace{\horspace} DenseNet121-24+SSD~\cite{cordone2022object}& 8.2 & 0.189 &-&-& 5  & 2.097 \\
         \hspace{\horspace} 64-ST-VGG+SSD~\cite{cordone2022performance} & 2.88 & 0.203 & - & - & \hspace{.3cm}1$ ^{(1)}$ & 1.557 \\
         \hspace{\horspace} \g{EMS-Res10-YOLO \cite{su2023deep}} $^{(2)}$ & \g{6.2} & \g{0.267}&\g{-} &\g{-}& \g{5} & \g{-}\\ 
         \hspace{\horspace} \g{EMS-Res18-YOLO \cite{su2023deep}} $^{(2)}$ & \g{9.3} & \g{0.286}&\g{-} & \g{-} & \g{5} & \hspace{.3cm}\g{0.393}$^{(3)}$\\ 
         \hspace{\horspace} Ours (no \gls{kd}) & 12.97 & 0.223 & 0.205 & 0.0119& 5 & 0.619\\
         \hspace{\horspace} Ours (with \gls{kd}) & 12.97 & 0.229 & 0.223 & 0.0043 & 5 & 0.999 \\ \hline
    \end{tabular}
    \\
    \flushleft{\footnotesize 
    \hspace*{.5cm}(1) Excludes preceding time steps which induce neuron membrane voltage.
    \hspace*{.5cm}(2) \etal{Su}~\cite{su2023deep} use non-spiking residual connections. \\
    \hspace*{.5cm}(3) Excludes energy consumption from the first coding layer.}}
    \label{tab:gen1_results}
\end{table*} 

\subsection{Implementation Details}
\subsubsection{Data} \label{subsubsec:data}
We train and evaluate all our models on the \textit{\gls{gen1} dataset} \cite{tournemire2020prophesee}. It consists of 39 hours of recordings with the 304$\times$240 pixel GEN1 sensor. Its event-based nature makes it particularly useful for training and evaluating \glspl{snn} as it natively provides spike-like and sparse input. \gls{gen1}  features an impressive number of 255,000 annotations for the two classes cars and pedestrians. These qualities made it an established benchmark for \gls{snn}-based object detectors \cite{cordone2022object, su2023deep}. Fig.~\ref{fig:inference} shows selected scenes.

Following the procedure of \cite{cordone2022object}, we sample 100\,ms of events preceding every annotation and split them into binary voxel cubes of 5 time steps, with each split into two micro time bins that are processed simultaneously. Together with the polarity of events, this gives us $2\times2 = 4$ input channels. However, as our non-spiking teacher model only uses one time step, we instead sample 20\,ms for its training to keep the information per time step similar.

\subsubsection{Hyperparameters} 
We train both our spiking and non-spiking models with the AdamW optimizer with a weight decay of 1e-4. 
While the non-spiking model uses a learning rate of 1e-3, the \gls{snn} models use 1e-4. All models use cosine annealing learning rate scheduler that reduces the learning rate to 1e-5. We clip our gradients at 1 to avoid exploding gradients. Due to faster convergence, the non-spiking model is trained for just 20 epochs while the spiking models are trained for 50 epochs. We initialized all but the output convolutions with the Kaiming Uniform method and zero bias. The heatmap head's last convolution's bias was initialized as -2.19 following \cite{zhou2019centernet} as it results in 1.0 after softmax activation. The size regression and offset heads' biases were initialized as 0.15 and 0.5 based on empiric convergence after some initial experiments. The corresponding convolution heads were initialized with normal distribution with standard deviation 1e-3.

\subsubsection{Testing}
Our main performance metric is the COCO \gls{map} \cite{lin2014microsoft} calculated over 10 IoU values ([.50:.05:.95]) as it is the de-facto standard metric for object detection. Unlike previous works \cite{cordone2022object}, we focus on the \textit{mean} performance of five trained model instances with different seeds as a more robust measurement and report the maximum, i.e., best performing model only for comparison.

Additionally, we aim to quantify the computational performance of our models as this is the main motivation behind the development of \glspl{snn}. In order to do so, we report the following metrics:
\begin{itemize}
    \item[$\bullet$] \textit{Number of parameters:} As in all neural networks, the number of parameters correlates with energy consumption. Additionally, embedded (neuromorphic) hardware as the desired deployment environment often features limitations on network size.
    \item[$\bullet$] \textit{\gls{ac} \& \gls{mac}:} We report both the \gls{ac} and \gls{mac} operations as measured by the SyOPs python library \cite{chen2023training} to calculate the theoretical energy needs for both the non-spiking and spiking models.
    \item[$\bullet$] \textit{Firing rate:} We also record the firing rate of the \glspl{snn} by calculating the proportion of active neurons (i.e., spikes) among all neurons (i.e., possible spikes) averaged over the test set and time steps (cf. \textit{sparsity} in \cite{cordone2022object}).
\end{itemize}

\begin{figure*}[htbp]
    \newcommand{\clsx}{-0.4}
    \newcommand{\imgoffsetx}{5.5}
    \newcommand{\imgoffsety}{4.4}
    \newcommand{\imgscale}{0.5}
    \centering
    \begin{tikzpicture}

        \node[anchor=south west,inner sep=0] at (0,0) {\includegraphics[scale=\imgscale]{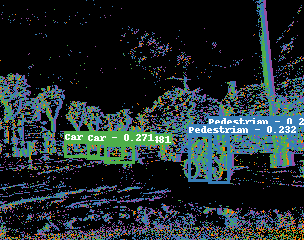}};
        \node[anchor=south west,inner sep=0] at  (0,\imgoffsety)  {\includegraphics[scale=\imgscale]{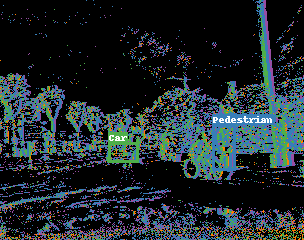}};
        
        \node[anchor=south west,inner sep=0] at (\imgoffsetx,0) {\includegraphics[scale=\imgscale]{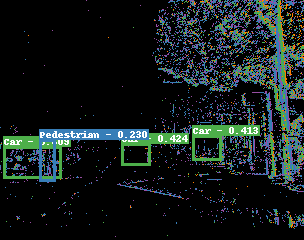}};
        \node[anchor=south west,inner sep=0] at (\imgoffsetx,\imgoffsety) {\includegraphics[scale=\imgscale]{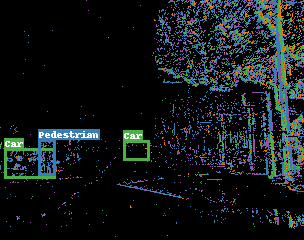}};
        
        \node[anchor=south west,inner sep=0] at (2*\imgoffsetx,0) {\includegraphics[scale=\imgscale]{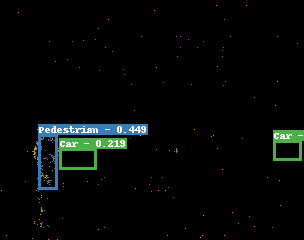}};
        \node[anchor=south west,inner sep=0] at (2*\imgoffsetx,\imgoffsety) {\includegraphics[scale=\imgscale]{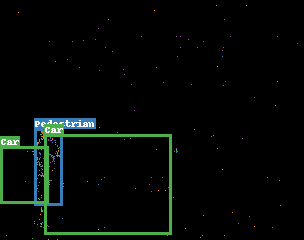}};

        \node[align=center, font=\footnotesize, rotate=90] at (\clsx, 6.2) {Ground Truth};
        \node[align=center, font=\footnotesize, rotate=90] at (\clsx, 2) {Prediction};

    \end{tikzpicture}
    
    \caption{Prediction of our best \gls{snn} model (bottom) and ground truth (top) for selected scenes of the \gls{gen1} dataset. The different pixel colors indicate the two micro time bins with each two polarities of brightness change, resulting in four input channels (cf. Section \ref{subsubsec:data}). Note that targets might be invisible if there is no camera or object motion.}
    \label{fig:inference}
\end{figure*}

To calculate the energy consumption of the \glspl{snn}, we mainly use the open-source tool \textit{syops-counter} provided by \cite{chen2023training}. As depthwise-separable convolutions and \gls{bn} layers after convolutions are useful for training, but introduce float values and thus \gls{mac} operations, we merge them according to  \cite{Rueckauer2017} before counting energy. We compute the energy consumption based on the validation split of the \gls{gen1} dataset. The results are reported in Table \ref{tab:gen1_results} (energy) and Table \ref{tab:num_ops} (\#operations, firing rate).

\subsection{Results}

\begin{table}[b]
 \centering
 \setlength{\tabcolsep}{1.9mm}
 \small
    \newcommand{\g}[1]{\textcolor{gray}{#1}}
    {
       
        \caption{Number of operations per  time step and firing rate.}
        \label{tab:num_ops}
        \begin{tabular}{l|c|c|c}
        \hline
        \textbf{\rule[-.125cm]{0mm}{.45cm}Model} & \textbf{MACs} & \textbf{ACs} & \textbf{Firing rate}\\ \hline \hline
             DenseNet121-24+SSD \cite{cordone2022object} &  0 & 2.33G & 37.20 $\%$ \\
             64-ST-VGG+SSD~\cite{cordone2022performance} & 0 & 1.73G & 38.87 \% \\
             \g{EMS-Res18-YOLO \cite{su2023deep}} & \g{-} & \g{-} & \g{20.09} $\%^{(1)}$\\ \hline
             Ours (\gls{ann} teacher) & 6.13G & 0.018G$^{(2)}$ & 100.0 $\%$ \\\hline
             Ours (no \gls{kd}) & 0 & 0.688G & 10.8 $\%$ \\
             Ours (with \gls{kd}) & 0 & 1.11G & 17.4 $\%$ \\ 
        \hline
        \end{tabular}
    }
    \flushleft{\footnotesize

    \hspace*{1.0cm}(1) Excludes energy consumption from the first coding layer. \\
    \hspace*{1.0cm}(2) Stem from binary event data input.}
\end{table} 

We report in Table~\ref{tab:gen1_results} results for three models: Our non-spiking \gls{ann} baseline (and teacher), the isomorphic \gls{snn} model trained without \gls{kd}, and the \gls{snn} model trained with \gls{kd} . Our results show that our simplified \gls{snn} model with \gls{kd} reaches a competitive \gls{map} of 0.229 (maximum) and 0.223 (mean), outperforming previous models in a similar offline setting \cite{cordone2022object} by a margin of 4\,\% \gls{map}. Our model also outperforms \cite{cordone2022performance}, which nominally uses one  time step to produce detections, but uses a semi-online setting with continuous neuron activity over sequences of 60s length, by 2.6\.\%. Fig.~\ref{fig:inference} shows object detections for selected scenes. 

The results in Table \ref{tab:gen1_results} also suggest that our \gls{kd} approach makes the \gls{snn} model consistently perform better and less reliant on outliers with a mean \gls{map} difference of +1.8\,\% and a standard deviation reduced by a factor of 2.7. Furthermore, we discover, that despite the higher number of parameters in our model, it is sparser and more energy efficient than comparable models (cf. Table \ref{tab:gen1_results} \& Table \ref{tab:num_ops}). However, we observe that the recent work of \cite{su2023deep}, who mix non-spiking structures into their partially spiking model, still performs better in terms of \gls{map} performance.

Regarding energy consumption, both our baseline \gls{snn} and \gls{kd}-based model significantly outperform both the 2.097\,mJ and 1.557\,mJ per time step of the models in \cite{cordone2022object} and \cite{cordone2022performance} with 0.619\,mJ and 0.999\,mJ respectively (see Table \ref{tab:gen1_results}).  Again, it must be noted that \cite{cordone2022performance} nominally uses only one time step (in comparison to our five), but in a different, continuous setup over 1200 sequential time steps.\cite{su2023deep} report a lower energy consumption, but do not count their initial convolutional layer's computational impact, making comparison with our results difficult. 

\subsection{Ablation Studies}
In order to explore the capabilities of our \gls{snn} models on working with fewer time steps, we first select the best instance of our models with and without \gls{kd} based on the results on the validation subset of the \gls{gen1} dataset. We then modify two parameters: The sequence length, i.e., the number of time intervals the input is divided into, and the sample window size, i.e., the time window in milliseconds we sample before each ground truth bounding box (see Section \ref{subsubsec:data}). We evaluate each 5-time-steps-trained model with a different number of time steps, ranging from 1 to 10. We do so twice: with a fixed time window of 100\,ms, i.e. the same information is compressed into fewer time steps of longer duration, and with a decreasing time window in which each time step has a fixed duration of 20\,ms. The results are shown in Fig.~\ref{fig:ablation}. We find that albeit performance unsurprisingly drops with the number of time steps, performance with 4 time steps still beats previous models with 5 time steps \cite{cordone2022object} and even 3 time steps still deliver decent performance. While simply dividing the same time window into more than 5 time steps does not improve performance significantly, a bigger time window does help slightly. However, past 140\,ms a bigger time window does not help either. 


\section{Discussion}
The main motivation behind our work is introducing a simple, well-performing architecture which strictly adheres to the definition of an \gls{snn}. Some previous works try to define complex, weighted signals as spikes while ignoring the additional computational cost these non-binary "spikes" introduce \cite{qu2023spiking}. Other works such as \cite{su2023deep} hide additional non-spiking operations within "spiking" blocks: Within their \textit{EMS-ResNet}, non-binary values are added, max-pooled and transmitted as residual connections over long distances, rather than cheap binary spikes. These not only incur additional costly \gls{mac} operations, it is also unclear whether spiking neuromorphic hardware will be able to support such complex neuron blocks. 

\begin{figure*}[ht]
\newcommand{\figscale}{0.5}
\newcommand{\imgscale}{0.212}

    \centering
    \hspace*{-2ex}\begin{subfigure}[h]{\figscale\textwidth}
        \includegraphics[trim={10 10 40 5}, clip, scale=\imgscale]{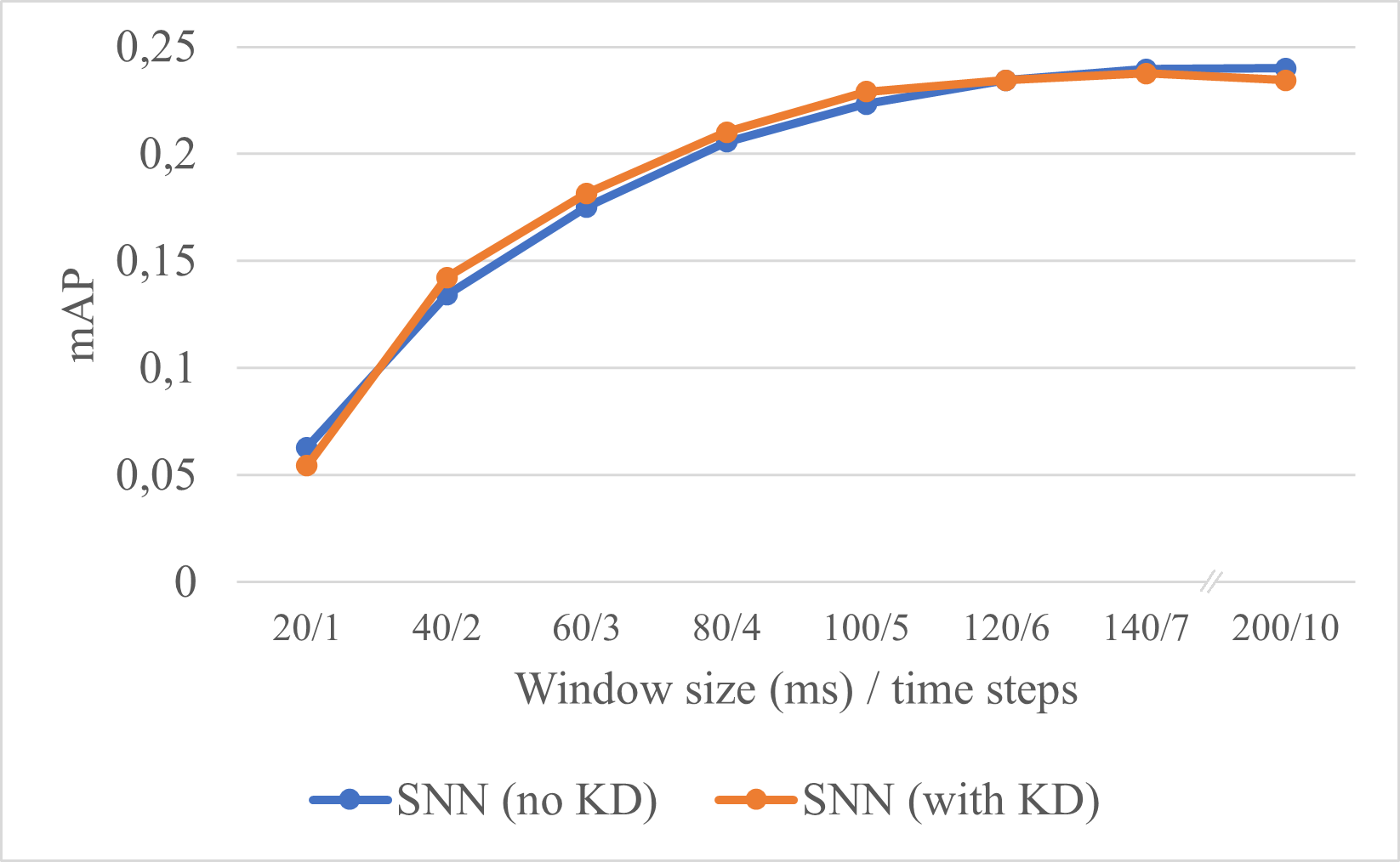}
        \caption{Fixed time window of 100\,ms}
        \label{fig:ablation_fix}
    \end{subfigure}
    \begin{subfigure}[h]{\figscale\textwidth}
        \includegraphics[trim={10 10 60 5}, clip,  scale=\imgscale]{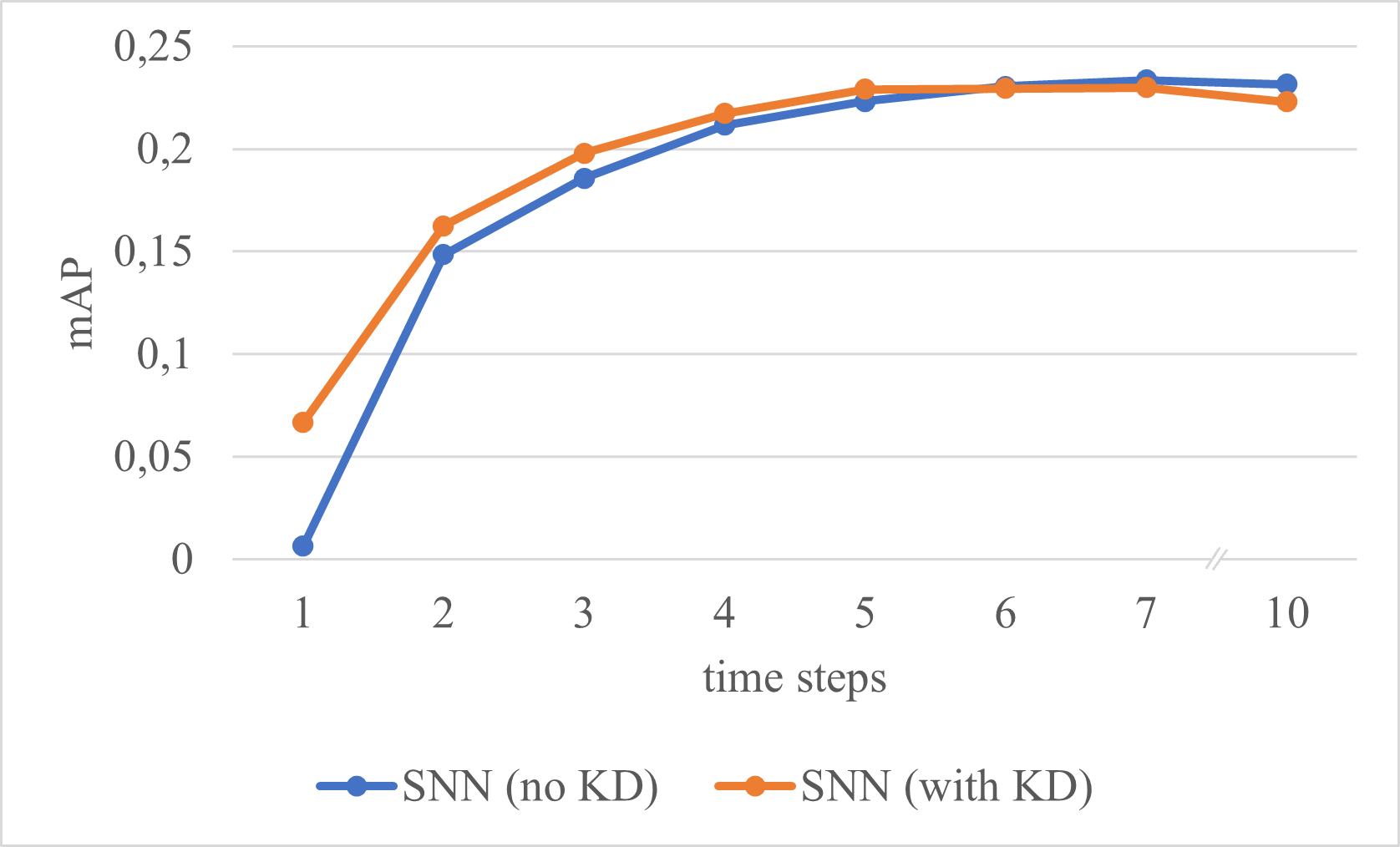}
        \caption{Variable time window of 20\,ms per step}
        \label{fig:ablation_var}
    \end{subfigure}
    \caption{Impact of the number of time steps in evaluation with a fixed (a) and variable (b) time window for sampling events. Shown is mAP of our best \gls{snn} models on the \gls{gen1} dataset \cite{tournemire2020prophesee}.}
    \label{fig:ablation}
\end{figure*}

\begin{figure*}[hb!] 
\newcommand{\figscale}{0.3}
\newcommand{\imgscale}{0.45}
\newcommand{\clsx}{-0.2}

\centering
\begin{tikzpicture}
    \node[anchor=south west,inner sep=0] at (0,0){
        \begin{subfigure}[h]{\figscale\textwidth}
                \includegraphics[trim={190 0 10 0}, clip, scale=\imgscale]{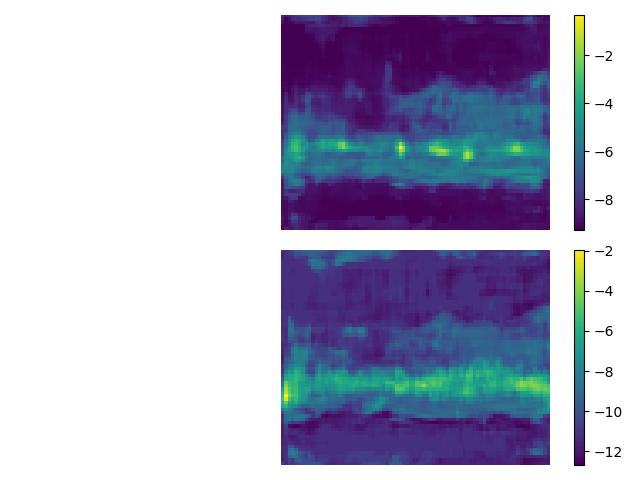}
            \caption{Non-spiking \gls{ann}}
        \end{subfigure}
        \begin{subfigure}[h]{\figscale\textwidth}
            \includegraphics[trim={195 0 10 0} , clip, scale=\imgscale]{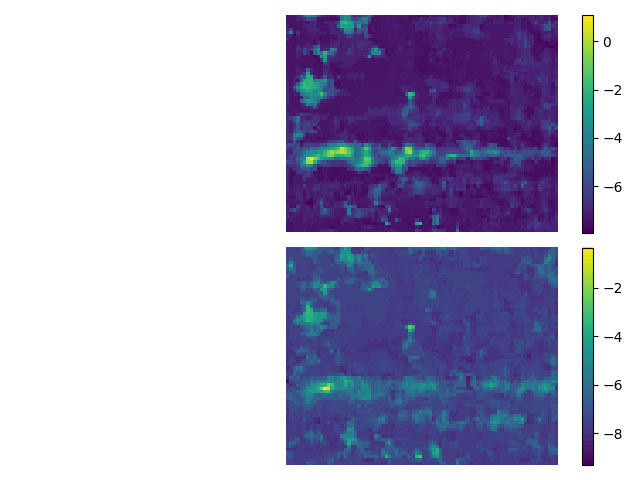}
                \caption{\gls{snn} (no \gls{kd})}
        \end{subfigure}
        \begin{subfigure}[h]{\figscale\textwidth}
            \includegraphics[trim={190 0 20 0} , clip, scale=\imgscale]{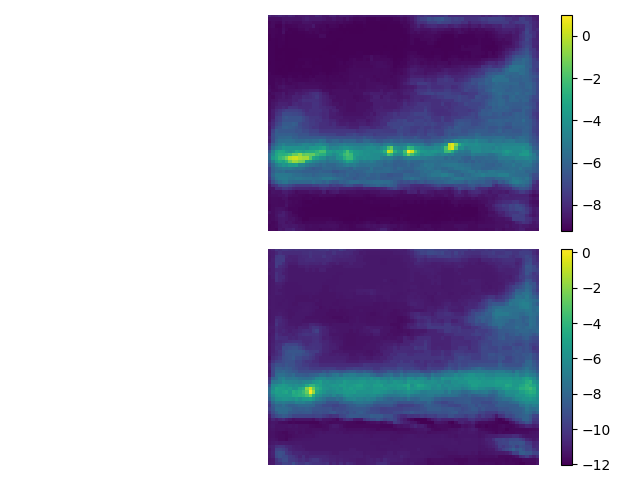}
                \caption{\gls{snn} with \gls{kd} }
        \end{subfigure}
    };
    \node[align=center, font=\small, rotate=90] at (\clsx, 2) {Pedestrian};
    \node[align=center, font=\small, rotate=90] at (\clsx, 4.5) {Car};
\end{tikzpicture}
\caption{Output of heatmap head (see Fig. \ref{fig:architecture}) averaged over time steps of the three evaluated models. Knowledge Distillation from the non-spiking \gls{ann} teacher to the \gls{snn} results in a less sparse, but smoother and ultimately better heatmap.}
\label{fig:hm_output}
\end{figure*}

In light of these concurrent works, it is our desire to keep our architecture as simple as possible and limit non-binary values to just the interface between convolutional and spiking activation layers as well as the final output, where the lack of subsequent neurons make spikes less valuable. Our M2U-Net-based \cite{laibacher2019m2unet} skip connections, connecting the encoder and decoder part of our backbone, merely transport sparse binary spikes. These are then concatenated and thus do not add \gls{mac} operations. We found that these skip connections are quite vital for gradient flow and enable the deep structure of the network; prior to adding them, the model would not learn at all on the \gls{gen1} dataset. Furthermore, our Spiking CenterNet is the first \gls{snn} detector without the expensive \gls{nms}, which also allows us to fully utilize the time dimension of the output. Lastly, Spiking CenterNet, due to its task-specific heads, is like the original CenterNet also easily expandable to other difficult tasks like 3D bounding box detection and pose estimation, which are not yet explored with \glspl{snn}.

We are also---to the best of our knowledge---the first to utilize Knowledge Distillation for training a spiking object detector. Since we observe that the best non-spiking \glspl{ann} still outperform \glspl{snn} by a wide margin (see Table \ref{tab:gen1_results}), it was our hope to transfer this performance to our \gls{snn} model. We observe that \gls{kd} indeed improves the result, especially mean performance. It can thus be used to make the training more consistent. We also observe that a \gls{kd}-boosted \gls{snn} seems to produce smoother, and according to \gls{map} better, heatmap outputs (see Fig.~\ref{fig:hm_output}). However, it also increases the number of spikes, therefore presenting a trade-off between performance and energy consumption.

Our evaluation reveals that despite our higher number of parameters, our \gls{snn} models actually use fewer spikes than comparable models (\cite{cordone2022object}, see Table \ref{tab:num_ops}). This results in a lower proportion of active neurons and thus lower firing rate. However, although a low firing rate is generally considered good \cite{cordone2022object}, it might also indicate an unnecessarily large network. Especially in light of possible limitations regarding neuron numbers in neuromorphic hardware, eliminating neurons which are inactive most or all of the time is logical. Nevertheless, first solving the object detection task at all to a satisfiable degree, which is a challenge in and of itself, is the prime priority of our work. 

Finally, our findings in evaluating our \gls{snn} models with fewer time steps indicate that our model can produce good results within a smaller time window than it has been trained for. This seems to confirm our decision of taking the mean of the overall network output over time as it makes the model less reliant on producing the correct output at all five time steps. During real-time inference, we expect the leaky membrane of the neurons to smooth over the asynchronous input events and a moving-average of the output could produce detections in each discrete output time step similarly to \cite{cordone2022performance}.
\section{Conclusion and Future Works}
We presented a new, versatile \gls{snn} architecture for object detection in the form of our {\em Spiking CenterNet}, consisting solely of simple building blocks and not requiring expensive \gls{nms}. Moreover, we are the first -- to the best of our knowledge --  to employ Knowledge Distillation for spiking object detection, which improves our baseline \gls{snn} model significantly. We observed that our \gls{snn} not only beats comparative previous work by 2.6\,mAP points, but also uses significantly less energy.
We demonstrate in our work that it is possible to push the performance of \glspl{snn} without stretching the definition of what constitutes an \gls{snn}. Furthermore, we show that even the simplest form of \gls{kd} can work for spiking object detection. More sophisticated \gls{kd} versions (e.g., for intermediate features) or more complex \gls{ann} teachers could be investigated in future works. We also plan to extend our approach to both RGB data input and also to different tasks such as 3D bounding box and human pose estimation, comparable to the original CenterNet \cite{zhou2019centernet}.

\bibliographystyle{ieeetr}

\begin{thebibliography}{10}

\bibitem{laibacher2019m2unet}
T.~Laibacher, T.~Weyde, and S.~Jalali, ``{M2U-Net}: Effective and efficient
  retinal vessel segmentation for real-world applications,'' in {\em IEEE/CVF
  Conference on Computer Vision and Pattern Recognition Workshops (CVPRW)},
  2019.

\bibitem{zhou2019centernet}
X.~Zhou, D.~Wang, and P.~Kr{\"a}henb{\"u}hl, ``Objects as points,'' {\em arXiv
  preprint arXiv:1904.07850}, 2019.

\bibitem{Gallego2020}
G.~Gallego, T.~Delbruck, G.~M. Orchard, C.~Bartolozzi, B.~Taba, A.~Censi,
  S.~Leutenegger, A.~Davison, J.~Conradt, K.~Daniilidis, and D.~Scaramuzza,
  ``Event-based vision: A survey,'' in {\em IEEE Transactions on Pattern
  Analysis and Machine Intelligence}, 2020.

\bibitem{hodgkin1952quantitative}
A.~L. Hodgkin and A.~F. Huxley, ``A quantitative description of membrane
  current and its application to conduction and excitation in nerve,'' {\em The
  Journal of Physiology}, vol.~117, no.~4, pp.~500--544, 1952.

\bibitem{delorme1999spikenet}
A.~Delorme, J.~Gautrais, R.~{van Rullen}, and S.~Thorpe, ``{SpikeNET}: A
  simulator for modeling large networks of integrate and fire neurons,'' in
  {\em Neurocomputing}, pp.~989--996, 1999.

\bibitem{fang2020incorporating}
W.~Fang, Z.~Yu, Y.~Chen, T.~Masquelier, T.~Huang, and Y.~Tian, ``Incorporating
  learnable membrane time constant to enhance learning of spiking neural
  networks,'' in {\em IEEE/CVF International Conference on Computer Vision
  (ICCV)}, 2021.

\bibitem{Pascanu2013}
R.~Pascanu, T.~Mikolov, and Y.~Bengio, ``On the difficulty of training
  recurrent neural networks,'' in {\em International Conference on Machine
  Learning (ICML)}, 2013.

\bibitem{shrestha2018slayer}
S.~B. Shrestha and G.~Orchard, ``Slayer: Spike layer error reassignment in
  time,'' in {\em Advances in Neural Information Processing Systems (NeurIPS)},
  2018.

\bibitem{neftci2019surrogate}
E.~O. Neftci, H.~Mostafa, and F.~Zenke, ``Surrogate gradient learning in
  spiking neural networks: Bringing the power of gradient-based optimization to
  spiking neural networks,'' {\em IEEE Signal Processing Magazine}, vol.~36,
  no.~6, pp.~51--63, 2019.

\bibitem{rumelhart1987learning}
D.~E. Rumelhart, G.~E. Hinton, and R.~J. Williams, ``Learning internal
  representations by error propagation,'' in {\em Parallel Distributed
  Processing: Explorations in the Microstructure of Cognition, Vol. 1:
  Foundations}, pp.~318--362, 1986.

\bibitem{kim2020spiking}
S.~Kim, S.~Park, B.~Na, and S.~Yoon, ``{Spiking-YOLO}: Spiking neural network
  for energy-efficient object detection,'' in {\em AAAI Conference on
  Artificial Intelligence}, 2020.

\bibitem{qu2023spiking}
J.~Qu, Z.~Gao, T.~Zhang, Y.~Lu, H.~Tang, and H.~Qiao, ``Spiking neural network
  for ultralow-latency and high-accurate object detection,'' {\em IEEE
  Transactions on Neural Networks and Learning Systems}, pp.~1--13, 2024.

\bibitem{cordone2022object}
L.~Cordone, B.~Miramond, and P.~Thi{\'e}rion, ``Object detection with spiking
  neural networks on automotive event data,'' in {\em International Joint
  Conference on Neural Networks (IJCNN)}, 2022.

\bibitem{cordone2022performance}
L.~Cordone, {\em Performance of spiking neural networks on event data for
  embedded automotive applications}.
\newblock PhD thesis, Universit{\'e} C{\^o}te d'Azur, 2022.

\bibitem{su2023deep}
Q.~Su, Y.~Chou, Y.~Hu, J.~Li, S.~Mei, Z.~Zhang, and G.~Li, ``Deep
  directly-trained spiking neural networks for object detection,'' in {\em
  Proceedings of the IEEE/CVF International Conference on Computer Vision},
  pp.~6555--6565, 2023.

\bibitem{hinton2015distilling}
G.~Hinton, O.~Vinyals, and J.~Dean, ``Distilling the knowledge in a neural
  network,'' in {\em Conference on Neural Information Processing Systems
  Workshops (NeurIPSW)}, 2015.

\bibitem{tran2022training}
T.~D. Tran, K.~T. Le, and A.~L.~T. Nguyen, ``Training low-latency deep spiking
  neural networks with knowledge distillation and batch normalization through
  time,'' in {\em International Conference on Computational Intelligence and
  Networks (CINE)}, 2022.

\bibitem{xu2023biologically}
Q.~Xu, Y.~Li, X.~Fang, J.~Shen, J.~K. Liu, H.~Tang, and G.~Pan, ``Biologically
  inspired structure learning with reverse knowledge distillation for spiking
  neural networks,'' {\em arXiv preprint arXiv:2304.09500}, 2023.

\bibitem{chen2017learning}
G.~Chen, W.~Choi, X.~Yu, T.~Han, and M.~Chandraker, ``Learning efficient object
  detection models with knowledge distillation,'' in {\em Advances in Neural
  Information Processing Systems}, vol.~30, Curran Associates, Inc., 2017.

\bibitem{tan2020efficientdet}
M.~Tan, R.~Pang, and Q.~V. Le, ``{EfficientDet}: Scalable and efficient object
  detection,'' in {\em IEEE/CVF Conference on Computer Vision and Pattern
  Recognition (CVPR)}, 2020.

\bibitem{bodla2017soft}
N.~Bodla, B.~Singh, R.~Chellappa, and L.~S. Davis, ``{Soft-NMS} -- {Improving}
  object detection with one line of code,'' in {\em IEEE International
  Conference on Computer Vision (ICCV)}, Oct 2017.

\bibitem{he2016deepb}
K.~He, X.~Zhang, S.~Ren, and J.~Sun, ``Deep residual learning for image
  recognition,'' in {\em IEEE Conference on Computer Vision and Pattern
  Recognition (CVPR)}, 2016.

\bibitem{spikingjelly}
W.~Fang, Y.~Chen, J.~Ding, Z.~Yu, T.~Masquelier, D.~Chen, L.~Huang, H.~Zhou,
  G.~Li, Y.~Tian, {\em et~al.}, ``{SpikingJelly}.''
  \url{https://github.com/fangwei123456/spikingjelly}, 2020.

\bibitem{zeiler2010deconvolutional}
M.~D. Zeiler, D.~Krishnan, G.~W. Taylor, and R.~Fergus, ``Deconvolutional
  networks,'' in {\em {IEEE} Conference on Computer Vision and Pattern
  Recognition (CVPR)}, 2010.

\bibitem{chen2023training}
G.~Chen, P.~Peng, G.~Li, and Y.~Tian, ``Training full spike neural networks via
  auxiliary accumulation pathway,'' {\em arXiv preprint arXiv:2301.11929},
  2023.

\bibitem{horowitz2014computing}
M.~Horowitz, ``1.1 computing's energy problem (and what we can do about it),''
  in {\em {IEEE} International Conference on Solid-State Circuits Conference
  ({ISSCC})}, 2014.

\bibitem{dampfhoffer2022snns}
M.~Dampfhoffer, T.~Mesquida, A.~Valentian, and L.~Anghel, ``Are snns really
  more energy-efficient than anns? an in-depth hardware-aware study,'' {\em
  IEEE Transactions on Emerging Topics in Computational Intelligence}, vol.~7,
  no.~3, pp.~731--741, 2022.

\bibitem{lemaire2022analytical}
E.~Lemaire, L.~Cordone, A.~Castagnetti, P.-E. Novac, J.~Courtois, and
  B.~Miramond, ``An analytical estimation of spiking neural networks energy
  efficiency,'' in {\em International Conference on Neural Information
  Processing}, pp.~574--587, Springer, 2022.

\bibitem{tournemire2020prophesee}
P.~De~Tournemire, D.~Nitti, E.~Perot, D.~Migliore, and A.~Sironi, ``A large
  scale event-based detection dataset for automotive,'' {\em arXiv preprint
  arXiv:2001.08499}, 2020.

\bibitem{hamaguchi2023hierarchical}
R.~Hamaguchi, Y.~Furukawa, M.~Onishi, and K.~Sakurada, ``Hierarchical neural
  memory network for low latency event processing,'' in {\em IEEE/CVF
  Conference on Computer Vision and Pattern Recognition (CVPR)}, 2023.

\bibitem{lin2014microsoft}
T.-Y. Lin, M.~Maire, S.~J. Belongie, J.~Hays, P.~Perona, D.~Ramanan,
  P.~Doll{\'a}r, and C.~L. Zitnick, ``Microsoft {COCO}: Common objects in
  context,'' in {\em European Conference on Computer Vision (ECCV)}, 2014.

\bibitem{Rueckauer2017}
B.~Rueckauer, I.-A. Lungu, Y.~Hu, M.~Pfeiffer, and S.-C. Liu, ``Conversion of
  continuous-valued deep networks to efficient event-driven networks for image
  classification,'' {\em Frontiers in Neuroscience}, vol.~11, p.~682, 2017.

\end{thebibliography}

\end{document}